\documentclass[journal]{journal}
\ifCLASSINFOpdf
  \usepackage[pdftex]{graphicx}

\usepackage{tikz}
\usepackage{comment}
\usepackage{amsmath,amssymb} 
\usepackage{color}

\usepackage[accsupp]{axessibility}  

\else
\fi
\begin{document}
%
\title{River Surface Patch-wise Detector Using Mixture Augmentation for Scum-cover-index}
%
%
%

\author{~Takato~Yasuno,~\IEEEmembership{}
           ~Junichiro~Fujii,~\IEEEmembership{}
           ~Masazumi~Amakata.~\IEEEmembership{}
\thanks{T. Yasuno is in charge of the Research Institute for Infrastructure Paradigm Shift (RIPPS), Yachiyo Engineering, Co.,Ltd., 5-20-8, Asakusabashi, Taito-ku, Tokyo, 111-8648, Japan (e-mail: tk-yasuno@yachiyo-eng.co.jp)}
\thanks{J. Fujii, M. Amakata are also in charge of RIIPS.}
\thanks{Manuscript received August 16, 2022; revised August 31, 2022.}}

%
%

\markboth{Journal of \LaTeX\ Class Files,~Vol.~6, No.~1, January~2007}%
{Shell \MakeLowercase{\textit{et al.}}: Bare Demo of IEEEtran.cls for Journals}
%



\maketitle
\thispagestyle{empty}

\begin{abstract}
Urban rivers provide a water environment that influences residential living. River surface monitoring has become crucial for making decisions about where to prioritize cleaning and when to automatically start the cleaning treatment. We focus on the organic mud, or ``scum", that accumulates on the river’s surface and contributes to the river’s odor and has external economic effects on the landscape. Because of its feature of a sparsely distributed and unstable pattern of organic shape, automating the monitoring process has proved difficult. We propose a patch-wise classification pipeline to detect scum features on the river surface using mixture image augmentation to increase the diversity between the scum floating on the river and the entangled background on the river surface reflected by nearby structures like buildings, bridges, poles, and barriers. 
Furthermore, we propose a scum-index cover on rivers to help monitor worse grade online, collect floating scum, and decide on chemical treatment policies. Finally, we demonstrate the application of our method on a time series dataset with frames every ten minutes recording river scum events over several days. We discuss the significance of our pipeline and its experimental findings.
\end{abstract}


%
\IEEEpeerreviewmaketitle

%
%
%
%

\section{Introduction}
\subsection{Related Works}
Automated garbage collection and river cleaning robots have been the center of river manager assisted robotics research for the past decade [1,2]. Additionally, river surface monitoring has become crucial to decide which areas are the worst and to automatically begin the cleaning process.
Since 1991, there have been numerous studies for understanding the scum formation mechanism, such as in-house experiments [3,4], field observations regarding organic sludge and odor [5,6], and automatic scum behavior monitoring using river surface computer vision [7,8]. 
Owing to the complex events and intertwined phenomena that include physical, chemical, biological, and hydrological features, it is not widely known how to fundamentally solve the river scum problem.

After a few days of rainfall, combined sewer overflows (CSOs) suddenly occur at the upstream of the river owing to scum and the highly dispersed residential living environment. This organic mud or ``scum" that appears on the river’s surface contributes to the river’s odor and has external economic effects on the landscape. 
Mizuta et al. [8] proposed a method for monitoring scum in an urban river’s tidal area. The 805 split lattice color images with a 20 $\times$ 20 pixel size were automatically used by the fixed point camera to determine whether or not scum was present. The accuracy for identifying the scum ranged from 58 to 88 percent using this straightforward neural network model with one hidden layer. The input variables included the 45 features of statistics and 25, 50, and 75 percentiles for each pixel’s RGB components. Additionally, they included input features like bridges and fences as reflected background features that were visible on the water surface. Although this was a preliminary version of the scum monitoring method, there is still room for accuracy enhancement and generalization to deep learning algorithms for pixel-by-pixel classification, also known as semantic segmentation, such as the U-Net [9].      
Nakatani et al. [10] studied detailed observations of scum using multiple cameras and sediment surveys in a tidal river to understand the generation and floating behavior of scum using the U-Net. They discovered evidence that the local flow may have a significant impact on the spatial-temporal behavior of the scum owing to the tide level. 
Semantic segmentation is unquestionably a possible option for a rough scum monitoring method.
If we use the marine debris object type [11,12], which includes bottles, cans, hooks, and tires, the U-Net with the ResNet34 backbone was still the best performing model. However, it has not always scored high accuracy with a mean Intersection of Union, 0.748.

However, the scum feature has a sparsely distributed and unstable pattern, which makes it difficult to improve the semantic segmentation accuracy.  
The scum segmentation on the river surface still has an over-prediction issue, and is insufficiently accurate for scum monitoring on the river water surface.     
The area of interest, the scum, was obscured by the water surface, which frequently reflected the nearby background features such as the building, bridge, pole, sign, sky, and trees.       
The authors propose a patch-wise classification method using image augmentation to improve the precise recognition of the river scum.

\subsection{Pixel-wise Segmentation vs Patch Classification}
First, we debate whether supervised learning such as that used in classification, object detection, and semantic segmentation, is a better deep learning approach. We can select semantic segmentation or classification if the target feature of the scum is not categorized into object context. It becomes challenging to annotate the scum feature on the pixel-wise region of interest (ROI) within the river surface in the case of the semantic segmentation. The scum features are sparsely dispersed, and the pattern is unstable and complex If we attempt to annotate one of the filled regions with multiple scum features, the trained learners would be low precision and the false negative error would increase resulting in a prediction of scum when it is actually background. Therefore, the authors proposed a patch-wise classification approach rather than the pixel-wise semantic segmentation. The patch size is 128 $\times$ 256.        
In contrast, we can consider an approach from unsupervised learning like the generator and anomaly detection such as the Variational Auto-Encoder (VAE) [13].

\begin{figure}[h]
\begin{center}
\includegraphics [width = 88mm] {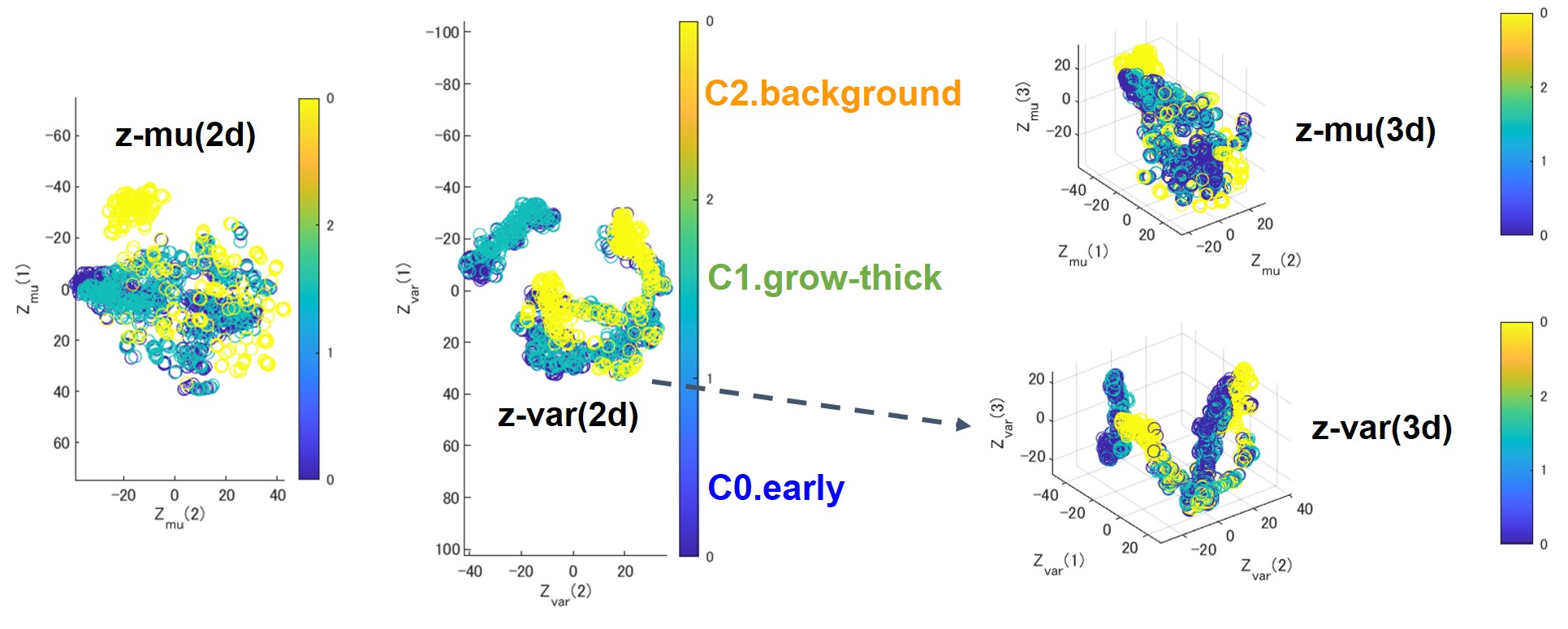}
\end{center}
\caption{Scum Feature of z-space Trained from Three Classes Conditional VAE (defined by C0:early scum, C1:grow-thick scum, C2:background. Training dataset has respectively (5,026, 4,070, 4,308). Test dataset has each classes (550, 450, 470).)}
\label{fig-1zvae}
\end{figure}

As illustrated in \textbf{Fig.1}, we demonstrated how to train the three classes of conditional VAE [14] using a dataset that was defined by C0. ``early scum,'' C1. ``grow-thick scum,'' and C2. ``background.'' Here, the bridge z-space contains 256 elements. Using t-SNE, we could dimension-reduce the z-mean and z-variance into plots of two and three dimensions. Hence, the three classes of river water surface, scum-generated feature, and a background that includes the building, train, bridge, barrier, pole, and tree, were not independent of each other. Here, a fake scene was reflected on the river’s water surface with a mirrored background rather than an actual image. River water surface images are very complex and are dependent on the water surface class and the target scum feature. Therefore, we did not select the generative learning and pixel-wise segmentation approaches. We discovered that it did not fit the dataset of river scum images for accurate scum detection.    
To tackle the low precision problem, this study proposed a patch-wise classification approach using mixture image augmentation.

\begin{figure}[h]
\begin{center}
\includegraphics [width = 88mm] {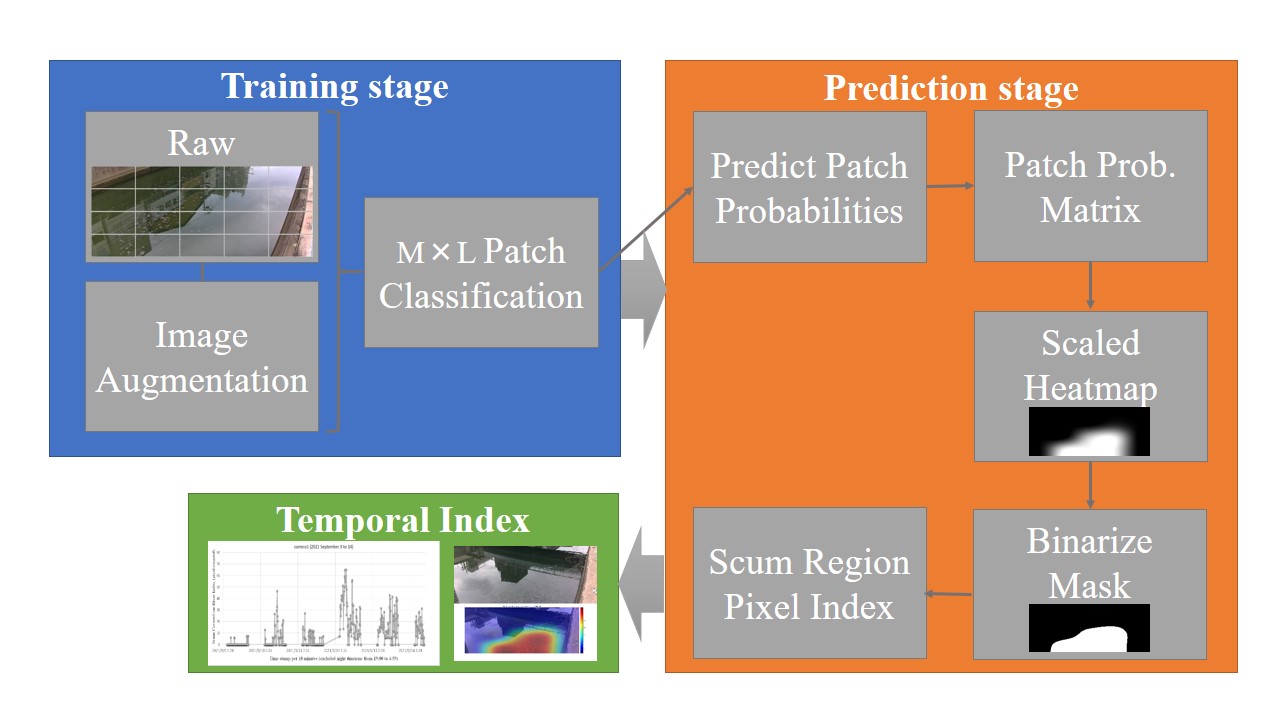}
\end{center}
\caption{Pipeline Overview from a Camera Frame Input to Scum-cover-index}
\label{fig-2pipe}
\end{figure}

\subsection{Overview of Pipeline}
As shown in \textbf{Fig.2}, we give an overview of the pipeline from the offline training stage to the online prediction and computing scum index. 
The first three components of the training stage are 1) data preparedness, 2) mixture image augmentation, 3) patch-classification deep learning. The data preparation process involves cropping a rectangle without the far region’s top and extracting four rows by five columns of patch images. 
Mixture image augmentation [15] transforms into diversified images using multiple raw images, such as the mixup [16,17], Cutout [18], and random image cropping and patching (RICAP) [19]. 
Convolutional neural networks (CNNs), such as ResNet50, ResNet101, and Inception-v3, are used in deep learning for patch classification. 
Second, the prediction stage using the pretrained deep network includes the following steps: 1) predict 20 patch probabilities; 2) combine their probabilities into a matrix; 3) create a scaled heatmap; 4) binarize the mask image; and 5) calculate the pixel count of the scum region.   
The prediction brings the total number of patch probabilities taken from the raw input frame to 20. 
Their probability values are combined into a matrix.
The matrix with four rows by five columns is transformed into a scaled heatmap with a grayscale intensity range of 0 to 255.
The heatmap can binarize a mask image with pixels set to 0 or 1.
The scum region pixel with a value of one on the mask image can be counted. 
Finally, the temporal scum index at a time stamp is computed as the value we call ``scum-cover-ratio" whose scum region pixel count is divided by the river region pixel count, ranging from 0 to 100 percent. 
Here, the river region can count pixels without a background. The background region is constant when the camera angle is fixed, allowing us to set the river region’s pixel count.   
We can repeatedly calculate the scum-cover-ratio using the raw input frame image collected every ten minutes. 
We can visualize the time series of scum-cover-ratio and draw the heatmap ranging from blue to red.

To distinguish between the floating scum feature and the entangled background on the river surface mirrored close to structures like buildings, bridges, poles, and barriers, we propose a patch classification pipeline to detect scum regions on the river surface using mixture augmentation. We also proposed a scum-cover-ratio index to help with online scum appearance degradation and decision-making regarding collecting floating scum and chemical treatment policy. Finally, we demonstrated how to use our pipeline on a time series of frames every ten minutes, recording the river scum vision for several days at an urban river in Japan 2021.

\section{River Surface Detection Method}
\subsection{Crop Far Region and Patch Classification}
The first three steps of the training stage are data preparation, mixture image augmentation, and deep learning patch classification. 
The data preparation operates to crop a rectangle without the top of the far region and to extract four rows by five columns of patch images. 
The size of the river vision camera used in this study is 1,280 pixels wide by 720 pixels high. We precisely cut the top of the rectangle at 1,280 pixels in width and 108 pixels in height, because of the extremely low resolution of this relatively remote area and the background noise from the bridge, building, pole, and barrier. 
The remaining image is therefore 128 $\times$ 256 in size and 512 by a width 1,280; hence, we can extract it based on a patch image. We can thus create a patch image with four rows by five columns.
We define the scum condition on the river surface: C0: early scum, C1: grow-thick scum, and C2: background.      
CNNs can be used to implement deep learning to classify patch images into three classes via supervised learning, e.g., ResNet18, ResNet50 [20], MobileNetv2 [21], and self-supervised and unsupervised learning [22]. 
We chose ResNet50, ResNet101, and Inception-v3 as practical CNNs for the supervised classification model; these deep networks are frequently used to facilitate transfer learning.

\subsection{Mixture Augmentation for Variety and Disentanglement}
The past decade has seen renewed importance of augmented image data for deep learning [13]. Image data augmentation is categorized into two i.e., basic image manipulation and deep learning approaches. The former augmentation includes 1) kernel filters, 2) geometric transformations, 3) random erasing, 4) mixing images, and 5) color space transformations. The latter consists of three components: 1) adversarial training, 2) neural style transfer, and 3) GAN model-based transformation. Particularly, geometric transformation, image blending, and neural style transfer have contributed to meta learning [15]. 
The deep learning-based augmentation could not be fitted to the target dataset of the river surface because of the features entangled between the region of interest in the scum and the background mirroring the building, bridge, pole, and barrier.    
This study proposes the fundamental image manipulations, such as mixing and random erasing of images. To detangle the river surface feature, the random erasing technique is straightforward and efficient as the regional dropout in the training images. We also believe that the augmentation of mixing images diversifies the river water surface feature. 
Using multiple raw images, the mixture image augmentation creates diversified images, like the mixup [16,17], Cutout [18], and RICAP [19]. CNN deep learning is used for patch classification.

As illustrated in \textbf{Fig.3}, the mixup is a linear combination manipulation using two randomly sampled images $X_i, X_j$. Here, we set the random weight parameter $\lambda \in [0,1]$. This weight parameter is randomly generated from the Beta distribution $Beta(\alpha,\alpha)$ where it takes a value between $[0,1]$. 
It is generated from the uniform distribution when $\alpha=1.0$. When $\alpha=0.2, 0.4$, the peakiness on both sides increases in a bustab-like shape. Therefore, we can write the following two equations to represent the mix-up image augmentation.
\begin{align}
  \tilde{M}      & = \lambda X_i + (1-\lambda) X_j \; \\
  \tilde{C}_{mixup} & = \lambda z_i + (1-\lambda) z_j
\end{align}
Here, two randomly selected images from different classes are labeled $z_i, z_j$, and the augmented new label $\tilde{C}_{mixup}$ results in one-hot encoding.
Notably, the mixup augmentation never enhances the overall performance of classification deep learning, even if two images were randomly selected from the same class.

\begin{figure}[h]
\begin{center}
\includegraphics [width =55mm] {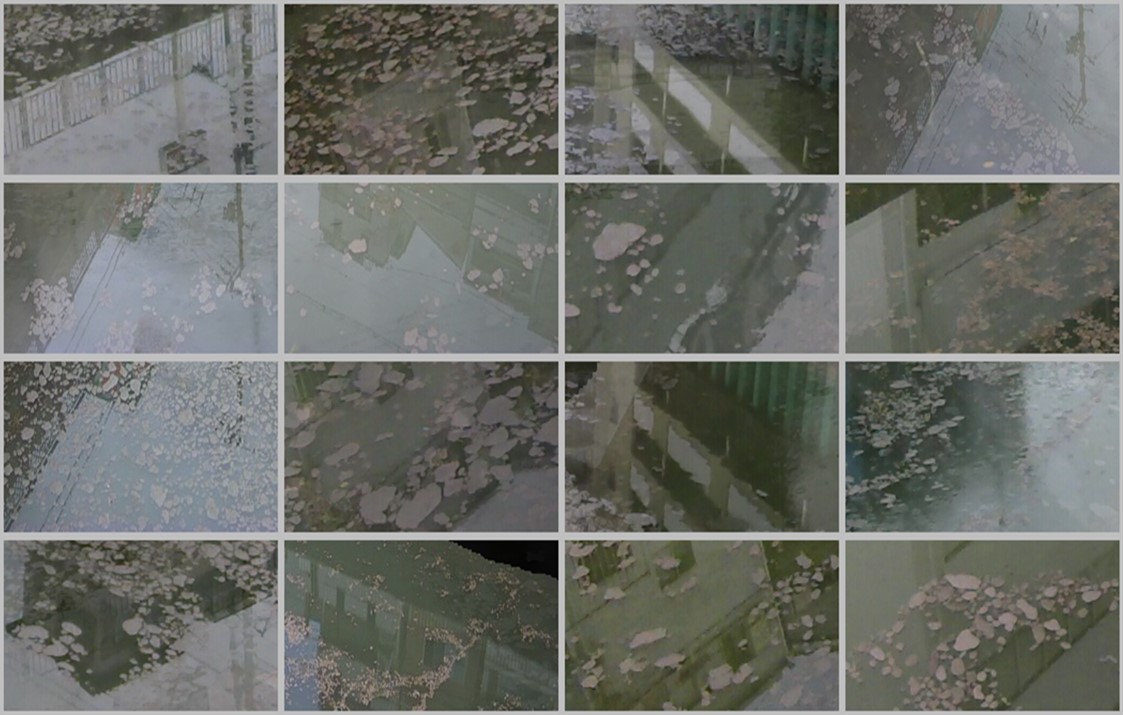}
\end{center}
\caption{Mixup Augmented Images of Scum grow-thick class}
\label{fig-3mup}
\end{figure}

As shown in \textbf{Fig.4}, the Cutout is a straightforward regularization technique that involves random masking out square input regions during training. Inspired by the dropout regularization mechanisms, this is an extremely easy implementation, but it could be combined with the existing form of data augmentation to further enhance model performance.
To apply to situations like occlusion, the Cutout image augmentation has been labeled {\it regional dropout}. This regional dropout has been used with the drop rate $d \in \{0.4, 0.5, 0.6\}$ in past experimental studies on natural image recognition datasets. 
We set the size of square mask images to one third of the total hight and width on input images. We randomly set the upper-left of corner on the mask within each training image.  
The complex entangled features on the river surface could be disentangled using the Cutout regularization. The entangled features included the scum floating organ and mirrored sky, bridge, building, and pole backgrounds.

\begin{figure}[h]
\begin{center}
\includegraphics [width =70mm] {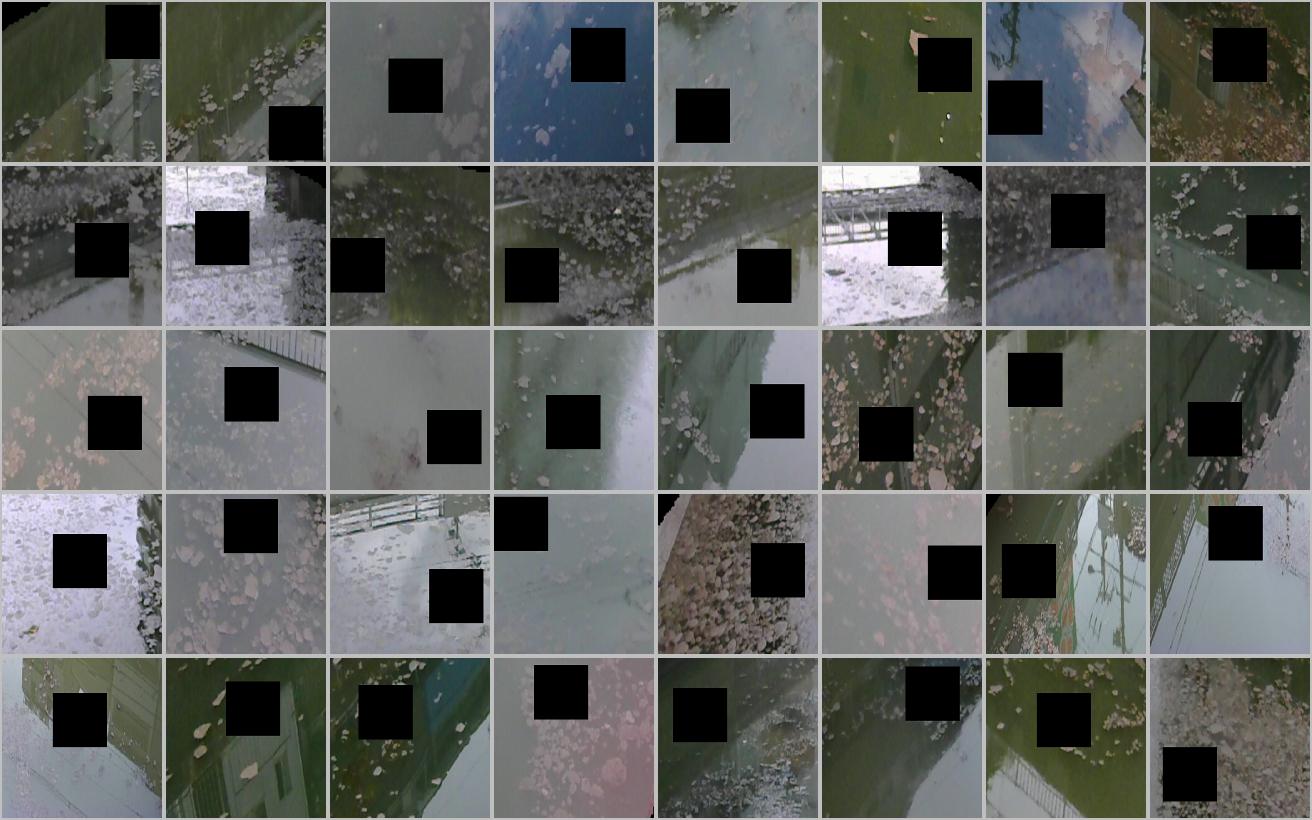}
\end{center}
\caption{Cutout Augmented Images of Scum grow-thick class}
\label{fig-4cut}
\end{figure}

As illustrated in \textbf{Fig.5}, RICAP involves mixing four randomly cropped images, respectively, and concatenating them into a new augmented image. RICAP greatly expands the variety of images and prevents overfitting of deep CNNs. 
Image data manipulation is done in three steps. 
First, we can randomly select four images $m\in{1,2,3,4}$ from the practice dataset. On the upper left $(m=1)$, upper right $(m=2)$, lower left $(m=3)$, and lower right sides $(m=4)$, we patch them in that order.
Second, we can crop the images separately.
$\bar{w}$ and $\bar{h}$ denotes the width and height of the training image, respectively.
We can randomly set the boundary position $(w,h)$ of the four images $m$ from a uniform distribution. This is known as the variant $anywhere$-RICAP.
\begin{align}
  w \sim U(0, \bar{w}), h \sim U(0, \bar{h}) \; 
\end{align}
Then we can automatically obtain the cropping sizes $(w_m,h_m)$ of the image $m$. i.e., $w_1=w_3=w$, $w_2=w_4=\bar{w}-w$, $h_1=h_2=h$, and $h_3=h_4=\bar{h}-h$. 
For cropping the four images $m$ following the sizes $(w_m, h_m)$, we can randomly determine the coordinates $(x_m, y_m)$ of the upper left corners of the cropped areas as $x_m \sim U(0, \bar{w} - w_m)$ and $u_m \sim U(0, \bar{h} - h_m)$.
Thirdly, we can patch the cropped images to construct a new image. 
We can mix the four images’ class labels with ratios proportional to the areas of the cropped images.   
Therefore, using the following equation, we define the target label $\tilde{C}_{RICAP}$ by mixing one-hot coded class labels $c_m$ of the four patched images with ratios $\Lambda_m$ proportional to their areas in the newly constructed image.

\begin{align}
  \tilde{C}_{RICAP}      & = \sum_{m\in \{1,2,3,4 \}} \Lambda_m c_m \; \\
            \Lambda_m    & = \frac{w_m h_m}{\bar{w}\bar{h}},
\end{align}
where $w_m h_m$ is the area of the cropped image $m$ and $\bar{w}\bar{h}$ is the area of the original image. 
Note that even if four images were randomly selected from the same class, the RICAP augmentation would never improve the overall classification performance. 

\begin{figure}[h]
\begin{center}
\includegraphics [width = 65mm] {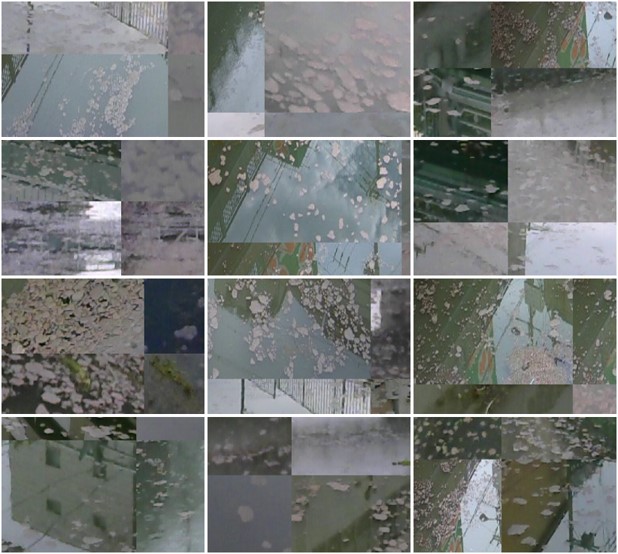}
\end{center}
\caption{RICAP-augmented Images of Scum grow-thick class}
\label{fig-5rcp}
\end{figure}

\subsection{Probability Scaled Heatmap and Scum-cover-Ratio}
In the prediction stage using the pretrained deep network, it includes the following steps: 1) predict 20 patch probabilities; 2) combine their probabilities into a matrix; 3) create a scaled heatmap; 4) binarize the mask image; and 5) pixel count of scum region.
The prediction brings the total number of patch probabilities extracted from the raw input frame to 20. 
Their probabilities’ values are combined into a matrix.
The matrix with four rows by five columns is transformed into a scaled heatmap with a grayscale intensity range of 0 to 255.
The heatmap can binarize a mask image with pixels set to 0 or 1.
The translated binary mask image and the gray scaled heatmap from the patch probability matrix are shown together in \textbf{Fig.6}.  

The scum region pixel whose value is one on the mask image can be counted. 
Finally, the temporal scum index at a time stamp is calculated as the value, we call ``scum-cover-ratio" whose scum region pixel count is divided by the river region pixel count, ranging from 0 to 100 percent.
\begin{align}
  ratio = \frac{pixels(scum)} {pixels(I - background)} = \frac{pixels(scum)} {pixels(river)} \; 
\end{align}
Here, $I$ denotes the cropped raw input image without the remote area of the upper rectangle. The river region can count the pixels without background. When the camera angle is fixed, the background area remains constant, allowing us to set the river region’s pixel count. 

\begin{figure}[h]
\begin{center}
\includegraphics [width = 70mm] {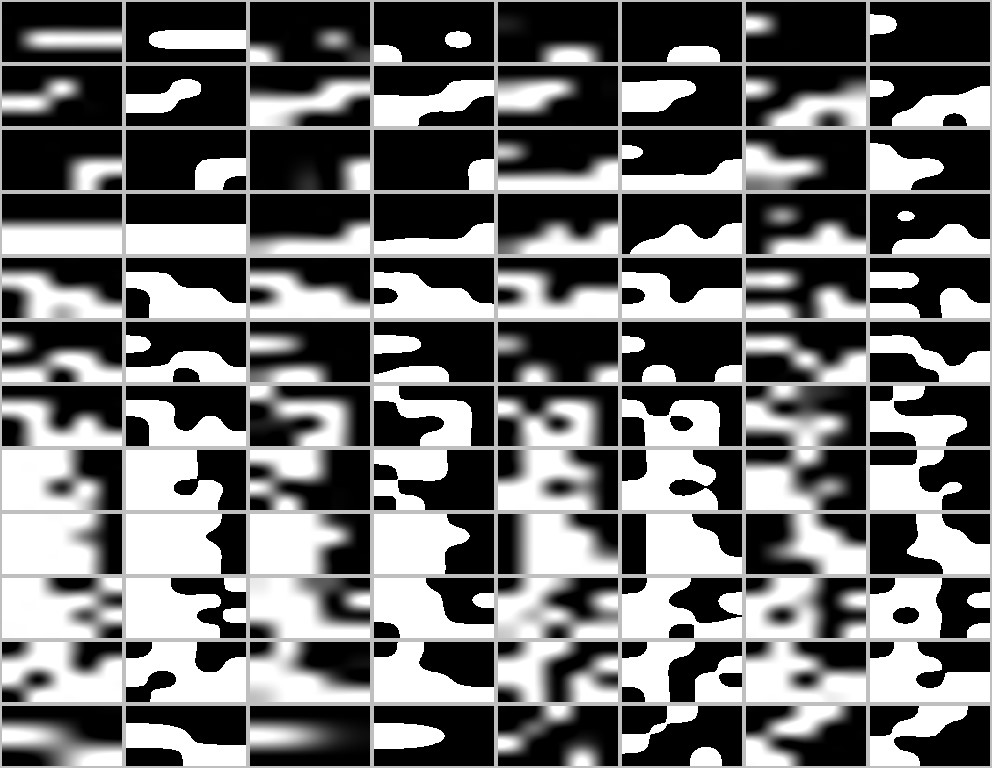}
\end{center}
\caption{Example of Binarized Masks from Patch Probability Scaled Heatmaps}
\label{fig-6}
\end{figure}

\section{Applied Results}
\subsection{Data Preparedness and Transfer Learning}
River scum is a rare occurrence, and it is difficult to collect images of its appearance before it gets too thick. We prepared a time series dataset to focus on the weeks when the scum first appeared, and we extracted frames per ten minutes from some recorded videos at an urban river around a densely populated area. A combined sewer station operates upstream of the city river.  
We demonstrated the application of our method on 5 day's time series with a frame every ten minutes, where river scum events have been occurring for several days.

For training classification, 884 frame images of scum occurrence in 2021 were used. The 13 cameras continuously recorded the scum conditions along an urban river and monitored the scum condition. We divided the training and test images in a 9:1 ratio. The image was 720 $\times$ 1,280 in both height and width.  
We have prepared 14,404 patches for the deep learning model’s training, and the 1,470 patches for evaluating test accuracy.

\subsection{Training Results and Test Accuracy}
We trained ResNet50, ResNet101, and Inception-v3 using a baseline model and several image augmentations such as mixup, Cutout, and RICAP. We implemented transfer learning using the Adam optimizer, which freezes 80 percent of feature extraction part of layers and removes the classifier part in the deep network. Then we add our new classifier softmax layer with the three classes for the scum dataset. We set the number of mini-batches to 64 and computed 20 epochs that iterates 225 per one epoch in total 4,500 iterations.

\begin{table}[h]
\caption{ResNet50: Test Accuracy Comparison of Patch Classification and Mixture Image Augmentation (3 classes defined by early scum, grow-thick scum, and background)} \label{table-1}
\begin{center}
\begin{tabular}{ c | c | c | c}
\hline 
\textbf{Augmentation} &\textbf{Test accuracy}  &\textbf{Precision}  &\textbf{  Recall  } \\
\hline 
\textbf{ResNet50 Baseline} & 97.4  & 97.6  & 94.6  \\
\hline 
+ mixup$(\alpha = 1.0)$  &  95.1  & 96.1   & 89.1  \\
\hline 
+ mixup$(\alpha = 0.2)$  &  97.4  & 97.6   & 94.4  \\
\hline 
+ mixup$(\alpha = 0.4)$  &  97.4  & 97.6   & 94.4  \\
\hline 
+ Cutout$(d = 0.4)$       &  97.8  &  99.5  & 94.0  \\
\hline 
+ Cutout$(d = 0.5)$       &  97.8  &  99.3  & 94.2  \\
\hline 
\textbf{+ Cutout}$(d = 0.6)$ & \textbf{98.0}  &  \textbf{99.3}  & 94.6 \\
\hline 
+ RICAP &  97.7  &  98.0   & \textbf{95.1}  \\
\hline 
\end{tabular}
\end{center}
\end{table}

The test accuracy comparison of the patch classification and mixture image augmentation is shown in \textbf{Table I}. Here, we define the three classes of river surface conditions, C0: early scum, C1: grow-thick scum, and C2: background.
The mixup augmentation never achieves a higher level of accuracy than the baseline ResNet50 model at the three cases. The linear mixing strategy was inefficient to generalize performance. 
However, when displaying the three rows of the Cutout augmentation, it outperformed the baseline model with a test accuracy of +0.6 and precision of +1.7. augmentation. The highest score among ablation studies was achieved by the drop rate $d = 0.6$. 
However, the recall accuracy was the same as in the baseline model. The RICAP augmentation +0.5 outperformed the recall accuracy of the baseline model.     
RICAP has improved the test accuracy and the precision over the baseline model by +0.3 and +0.4, respectively. 
For scum detection and monitoring to decrease the scum-positive false error, where predictions negative scum are actually scum, the recall accuracy is essential.  
Therefore, we consider the results provided by RICAP as practically the best among these ResNet50 + augmentation studies. 

\begin{table}[h]
\caption{ResNet101: Test Accuracy Comparison of Patch Classification and Mixture Image Augmentation} \label{table-2}
\begin{center}
\begin{tabular}{ c | c | c | c}
\hline 
\textbf{Augmentation} &\textbf{Test accuracy}  &\textbf{Precision}  &\textbf{  Recall  } \\
\hline 
\textbf{ResNet101 Baseline} & 96.7  & 96.5  & 93.6  \\
\hline 
+ mixup$(\alpha = 1.0)$  &  96.5  & 98.2   & 91.2  \\
\hline 
+ mixup$(\alpha = 0.2)$  &  96.3  & 98.1   & 90.8  \\
\hline 
+ mixup$(\alpha = 0.4)$  &  96.2  & 97.9   & 90.8  \\
\hline 
+ Cutout$(d = 0.4)$       &  97.8  &  99.3  & \textbf{94.9} \\
\hline 
+ Cutout$(d = 0.5)$       &  97.6  &  \textbf{99.9}  & 92.7  \\
\hline 
\textbf{+ Cutout}$(d = 0.6)$ & \textbf{98.2}  &  99.8  & \textbf{94.9} \\
\hline 
+ RICAP &  97.0  & 98.9   & 92.7  \\
\hline 
\end{tabular}
\end{center}
\end{table}

As shown in \textbf{Table II}, at the three rows of the mixup augmentation, it achieved a higher precision of accuracy than the baseline ResNet101 model. However, the recall accuracy never improved.
When displaying the three rows of the Cutout augmentation, it outperformed the baseline model with a test accuracy of +1.5, precision of +3.3, and recall of +1.3.
The highest score among ablation studies was achieved by the drop rate $d = 0.6$. 
The RICAP augmentation improved the test accuracy and precision over the baseline model by +0.3 and +2.4, respectively. However, the recall never improved.
Therefore, the Cutout($d=0.6$) is the best result among these ResNet101 + augmentation studies. 

\begin{table}[h]
\caption{Inceptionv3: Test Accuracy Comparison of Patch Classification and Mixture Image Augmentation} \label{table-3}
\begin{center}
\begin{tabular}{ c | c | c | c}
\hline 
\textbf{Augmentation} &\textbf{Test accuracy}  &\textbf{Precision}  &\textbf{  Recall  } \\
\hline 
\textbf{Inception-v3 Baseline} & 96.7  & 98.8  & 91.2  \\
\hline 
+ mixup$(\alpha = 1.0)$  &  96.1  & 98.1   & 90.1  \\
\hline 
+ mixup$(\alpha = 0.2)$  &  95.5  & 98.1   & 88.9  \\
\hline 
+ mixup$(\alpha = 0.4)$  &  95.9  & 98.1   & 89.7  \\
\hline 
\textbf{+ Cutout}$(d = 0.4)$ & \textbf{97.6} & \textbf{99.5} & \textbf{94.4}  \\
\hline 
 + Cutout$(d = 0.5)$       &  97.2  &  99.3  & 92.5  \\
\hline 
+ Cutout$(d = 0.6)$ & 97.1  & 98.4  & 93.4 \\
\hline 
+ RICAP & 96.8  & 97.9 & 91.9 \\
\hline 
\end{tabular}
\end{center}
\end{table}

As shown in \textbf{Table III}, the mixup augmentation did not improve accuracies other than that of the baseline Inception-v3 model. 
When displaying the three rows of the Cutout augmentation, it outperformed the baseline model with a test accuracy of +0.9, precision of +0.7, and recall of +3.2.
The highest score among ablation studies was achieved by the drop rate $d = 0.4$. 
The RICAP augmentation improved the test accuracy and recall over the baseline model by +0.1 and +0.7, respectively. However, the precision never improved.
Therefore, the Cutout($d=0.4$) is the best result among these Inception-v3+ augmentation studies. 

Thus, we chose the ResNet101 + Cutout($d=0.6$) result using the regional dropout of augmentation strategy from the three model's ablation studies, setting the pretrained network to predict the patch probabilities and computing the scum-cover-ratio.

\begin{figure}[h]
\begin{center}
\includegraphics [width = 88mm] {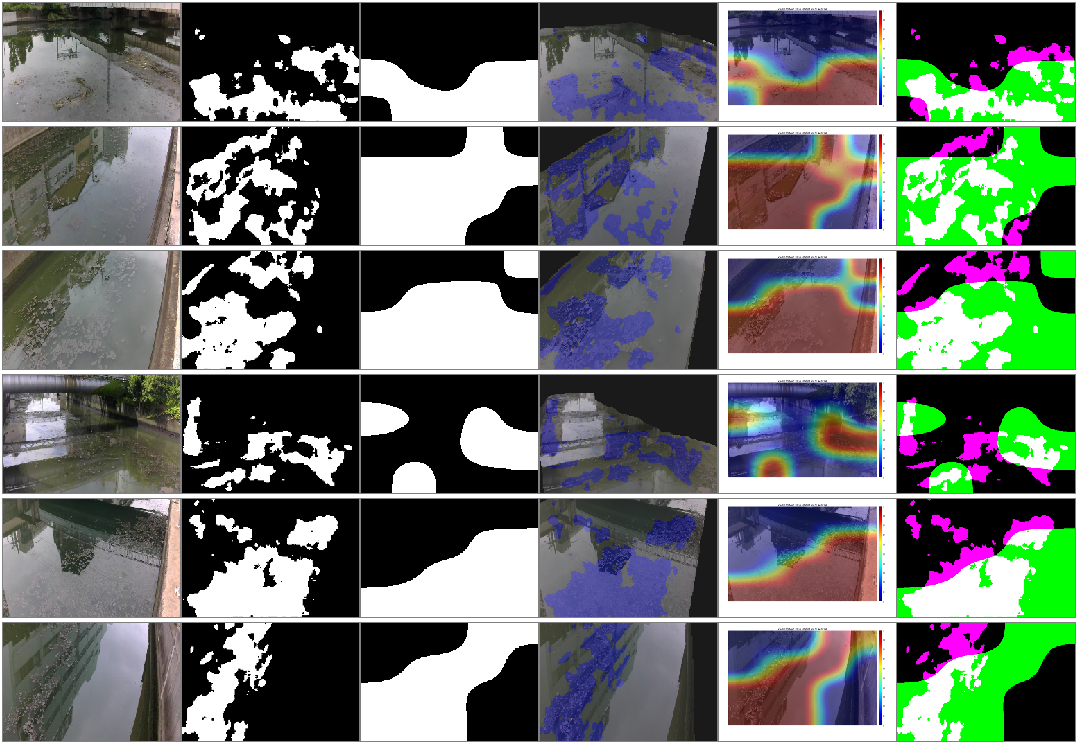}
\end{center}
\caption{Comparison of U-Net Segmentation and Patch Classification(in the right, raw image, U-Net prediction, scum-cover output(ours), segmentation overlaid, scaled heatmap(ours), fusion of two predictions)}
\label{fig-7QA}
\end{figure}

\begin{figure}[h]
\begin{center}
\includegraphics [width = 88mm] {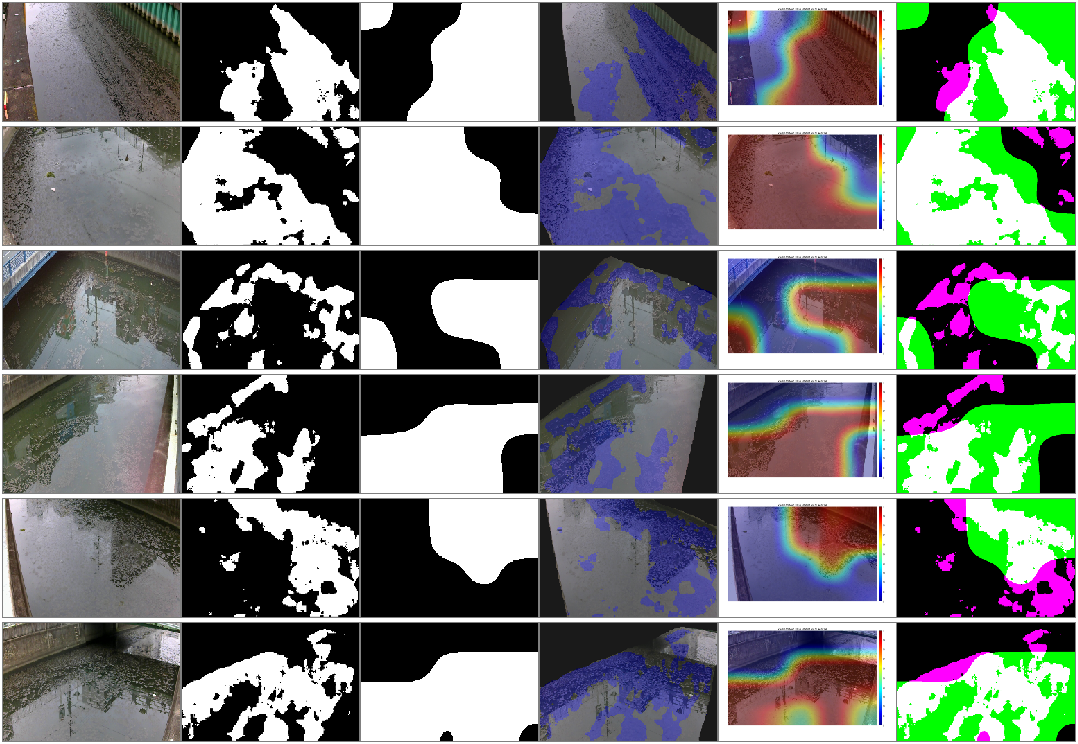}
\end{center}
\caption{Comparison of U-Net Segmentation and Patch Classification(in the right, raw image, U-Net prediction, scum-cover output(ours), segmentation overlaid, scaled heatmap(ours), fusion of two predictions)}
\label{fig-8QA}
\end{figure}

\subsection{Qualitative Analysis and U-Net Similarity}
We demonstrated the application of our patch-wise detector method and existing pixel-wise segmentation using 884 images from 13 cameras monitoring a clean river.  
As depicted in \textbf{Fig.7} and \textbf{Fig.8}, compared to the state-of-the-art of semantic segmentation approach using U-Net, we visualized a montage of examples including 6 columns per camera on the river from 12 cameras. It is depicted on the right as follows, 1) raw image cropped the upper background, 2) U-Net prediction output, 3) scum-cover prediction output(ours), 4) semantic segmentation overlaid, 5) scaled heatmap based on the patch probabilities of scum class (ours), 6) fusion of two predictions comparing the U-Net and our patch-wise detector.
Using a U-Net with 46 layers and a dice loss function, we annotated 200 images randomly selected from one camera H5 and trained 100 epochs in ten hours. The accuracy scores are mIoU 0.785, scum class IoU 0.613, and background class IoU 0.958 as a reference study.
Thus, the patch classification prediction output approximately covered the pixel-wise segmentation output at each camera to the first order. The white region indicates that two methods' outputs are completely matched at the six columns of fusion. Green represents a lower level of precision and magenta represents a lower level of recall accuracy.
\begin{figure}[h]
\begin{center}
\includegraphics [width = 88mm] {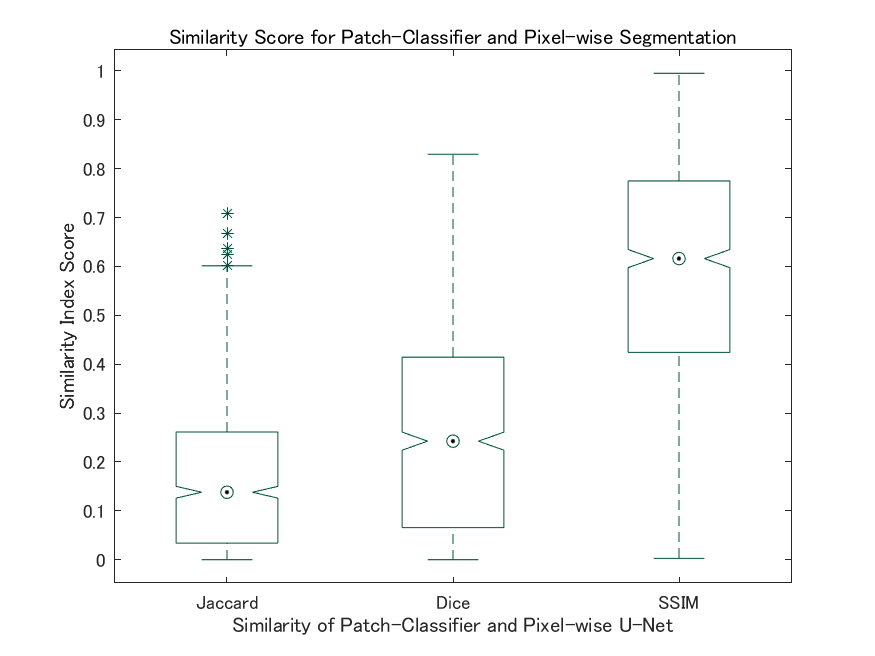}
\end{center}
\caption{Similarity Score of the U-Net Segmentation and Our Patch Classification (boxplots in the right, Jaccard similarity, Dice similarity, Structural Similarity(SSIM))}
\label{fig-9simBox}
\end{figure}
      
\textbf{Fig.9} shows the similarity score of the pixel-wise segmentation and our patch classification, whose boxplots are on the right, Jaccard similarity, Dice similarity, Structural Similarity(SSIM). The median scores are 0.138, 0.242, 0.616 respectively, the 75 percent quantile values are 0.261, 0.414, 0.775 and the 25 percent quantile scores are 0.033, 0.065, 0.424 respectively. 
Although, the patch-wise classification did not completely match the pixel-wise segmentation method, the patch classification predictor was first order approximately similar to the pixel-wise segmentation in 61.6 percent as the SSIM similarity on the median score.

\subsection{Visualize Temporal Scum-cover-Ratio for Monitoring}
As shown in \textbf{Fig.10} and \textbf{Fig.11}, the peaks of the scum-cover-ratio could precisely detect the peak of scum appearance. This refers to the feasibility of using our proposed scum-cover-ratio index to monitor the grade of scum generation. 
In contrast, the building background noise reflected on the surface of the river has, however, sometimes been over-predicted.
For example, the waving on the river surface after a boat has passed causes scum-negative false errors. 
Furthermore, it sometimes made an incorrect prediction regarding the light rainfall noise reflected on the river surface. This explains why the scum feature from the AI view of the pretrained classification network and the reflecting rainfall on rivers are similar. Because the training dataset of this work excluded the rainfall background to generalize the hydrologic context.

\begin{figure}[h]
\begin{center}
\includegraphics [width = 88mm] {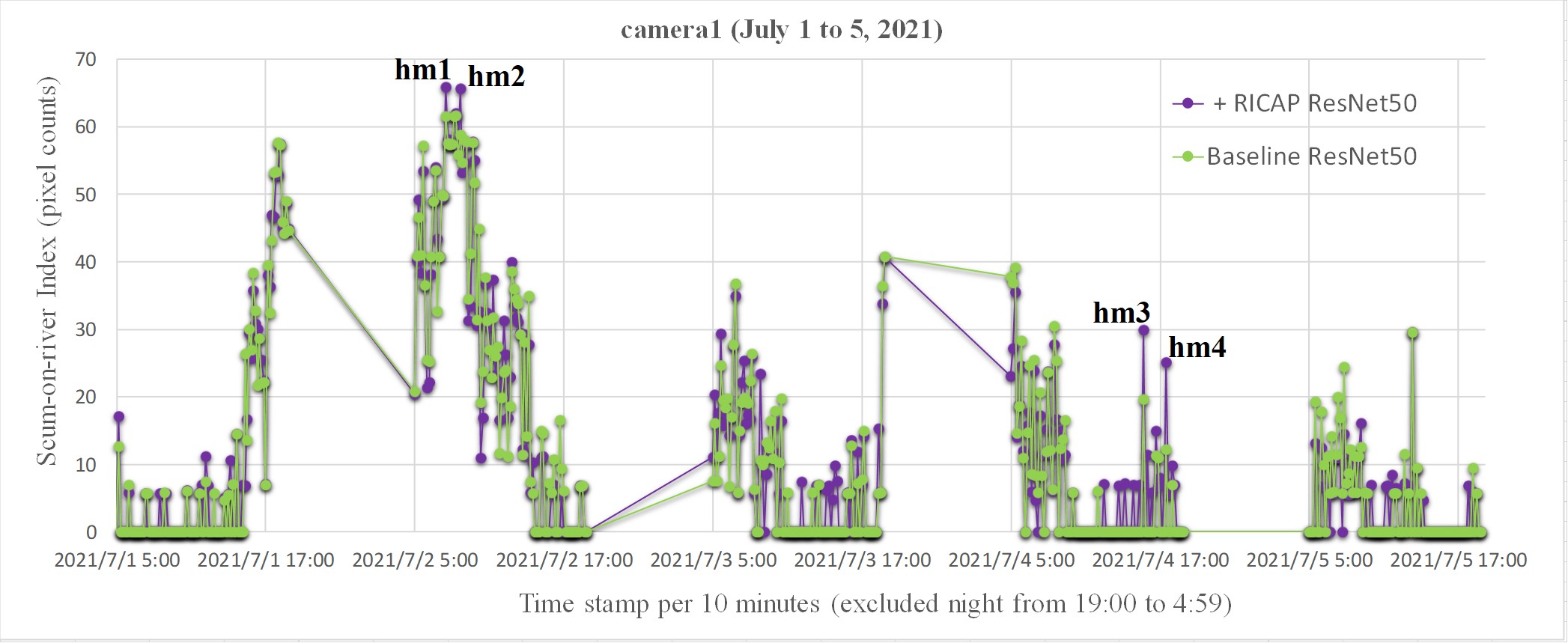}
\end{center}
\caption{Results of Temporal Scum-cover-ratio Index (camera1)}
\label{fig-10cam1idx}
\end{figure}

\begin{figure}[h]
\begin{center}
\includegraphics [width = 88mm] {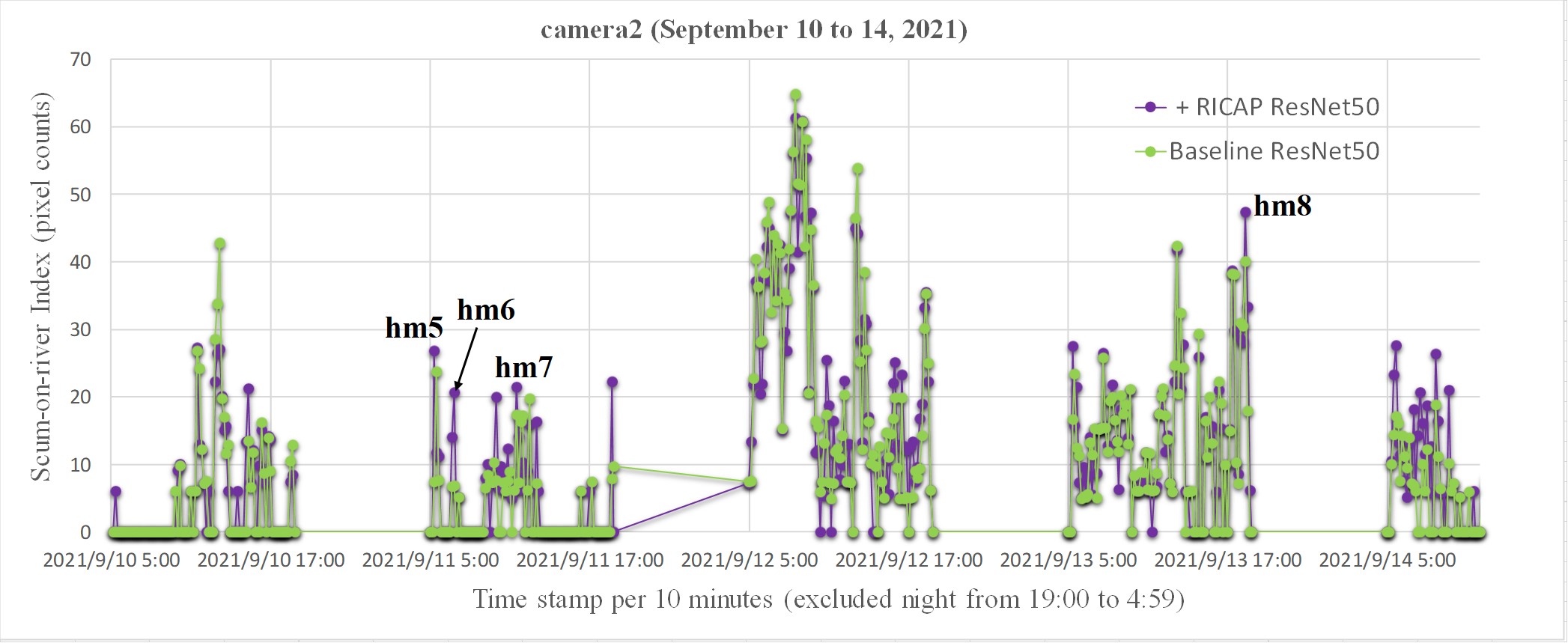}
\end{center}
\caption{Results of Temporal Scum-cover-ratio Index (camera2)}
\label{fig-11cam1idx}
\end{figure}

In \textbf{Fig.12}, a method for computing scum-cover-ratio for a monitoring camera H5 observing river surface is shown. This plot shows the temporal scum-cover-ratios for model comparison between the ResNet50, ResNet101, Inception-v3 with mixture augmentation. The performance of the augmented ResNet50 model sometimes sky rocketed. 
Therefore, the augmented ResNet101 model has exhibited higher stability than other models.

\begin{figure}[h]
\begin{center}
\includegraphics [width = 88mm] {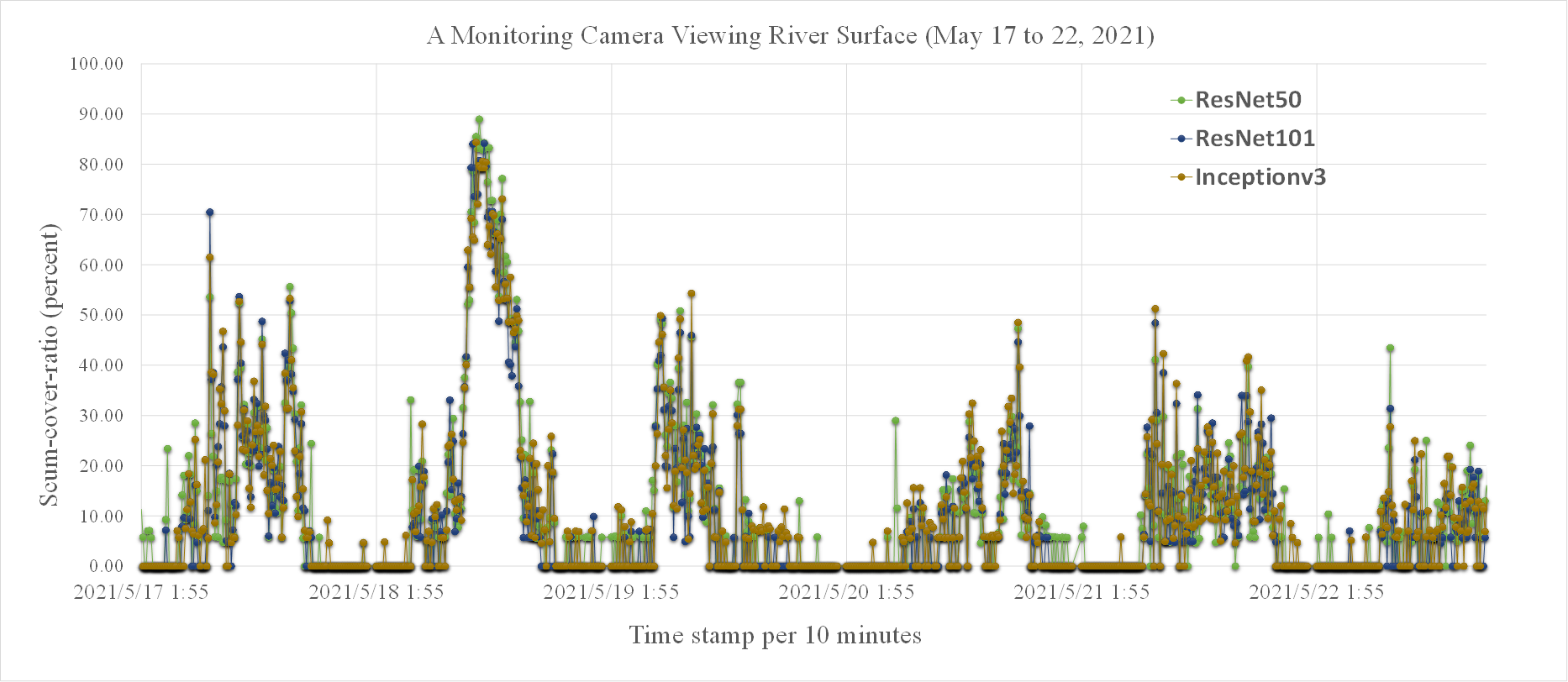}
\end{center}
\caption{Results of Temporal Scum-cover-ratio Index (model comparison, ResNet50, ResNet101, Inception-v3)}
\label{fig-12cam1idx}
\end{figure}

\section{Concluding Remark}
\subsection{Experimental Studies and Lessons}
We suggested using image augmentation to increase the variety and disentangled regularization in a patch-wise detector to identify scum features on river surfaces.
We precisely trained the practical deep network’s baseline. We have enhanced river’s surface images using the mixup, Cutout, and RICAP for accuracy comparison. 
We found that ResNet101 + Cutout($d=0.6$) outperformed the baseline model and other augmentations in terms of recall accuracy, 94.9 percent, and precision, 99.8 percent. We also discovered that the mixture strategy could improve the best recall by 95.1 percent in the ResNet50 + RICAP augmentation.
Furthermore, we provided an assisted metric to make it possible to compute a ``scum-cover-ratio" index for river scum monitoring and heatmap visualization.
With the grow-thick scum class, we could set the average probability’s bottom line to 0.01, insignificant probability that prevented an unstable scale heatmap. Using this setting, we could illustrate a scum-cover-ratio of greater than five percent when we implemented experimental studies to our river surface dataset.   
Finally, we demonstrated how to use our pipeline on images from urban cameras. The dataset of time series frames every ten minutes included scum growing events and thick conditions on two weeks, May, July, and September 2021 in Japan. 
We found that the limitation that background noise such as building reflections on the surface, and light rain waves on the river’s surface influenced the false positive error. We also discovered the feasibility of detecting the scum appearance in the early morning hours.
Furthermore, we demonstrated the application of our patch-wise detector and the pixel-wise segmentation approach using the U-Net on the dataset of 13 cameras monitoring a clean river. Our patch-wise detector was first order approximately similar to the pixel-wise segmentation in 61.6 percent according to the structural similarity on the median score.

\subsection{Future Works}
The off-line training and prediction results are limited in this study. Several obstacles for on-line assisted river managers address the water surface vision problem. 
First, it reduces rain noise in relation to rain reflection on the river surface on rainy days. Second, we should monitor the river surface during the night, in the early morning five o'clock and after 19 o'clock because the scum growing event could have occurred at night. Third, it becomes crucial to forecast the scum excess time more than the water quality level. We can collect additional time series datasets such as temperature, rainfall, and river water level. Using these temporal variables, we can use multi-mode learning to forecast the growing scum trend and the peak of the scum index level.
Fourth, the rare frequency of scum events makes the supervised training  time consuming and results in high data collection cost. Therefore, we utilize both self-supervised and unsupervised approaches to create more general pipelines.  
River water environment will still impact urban life around the world.   
We attempt to automate river surface monitoring that assists for water environment cleaning.


%

\appendices

\section{Dataset and Class Definition}
\subsection{Training Image Examples for Patch Classification}
The authors prepared a river-scum image dataset with a total number of 14,404 images for training. Moreover, we prepared a second unseen test dataset consisting of 1,470 images. The test image dataset contained 533, 467, and 470 images for each class C0, C1, and C2, respectively. Here, we illustrate the examples of each class patch image that is randomly sampled 40 times. The size of each patch image is 128$\times$256 in height and width.

As illustrated in \textbf{Fig.13}, we show the C0 class of images for classification training that consists of 5,576 images in the C0 class. Here, we randomly sampled 40 patch images from the C0 early scum class of the dataset.
The C0 class has been the domain of zero scum on the river surface, or the small scum on the river initially whose mixture were zero scum and extremely small scum on the river.   

\begin{figure}[h]
\begin{center}
\includegraphics [width = 88mm] {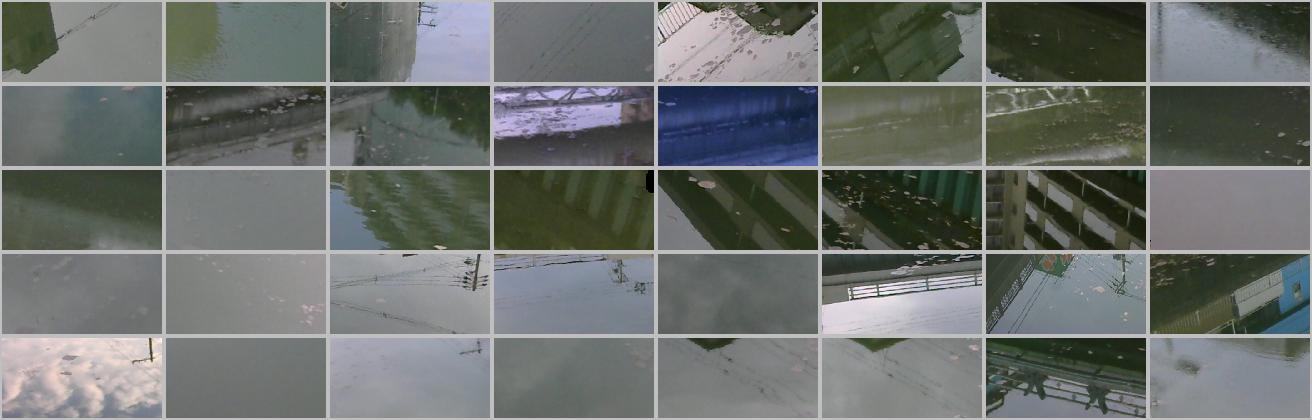}
\end{center}
\caption{Example of Class \textbf{``C0 early scum''} Patch Images}
\label{fig-m1C0}
\end{figure}

In addition, as illustrated in \textbf{Fig.14}, we show the C1 class of images for classification training which consists of 4,520 images. We randomly sampled 40 patch images from the C1 grow-thick scum class. The C1 class was the domain whose scum was growing, or with thick scum on the river surface reflecting these neighbor structures such as bridges, buildings, barriers, and poles. On a sunny day, the sky and clouds have been reflected of the scum growing on the river surface.

\begin{figure}[h]
\begin{center}
\includegraphics [width = 88mm] {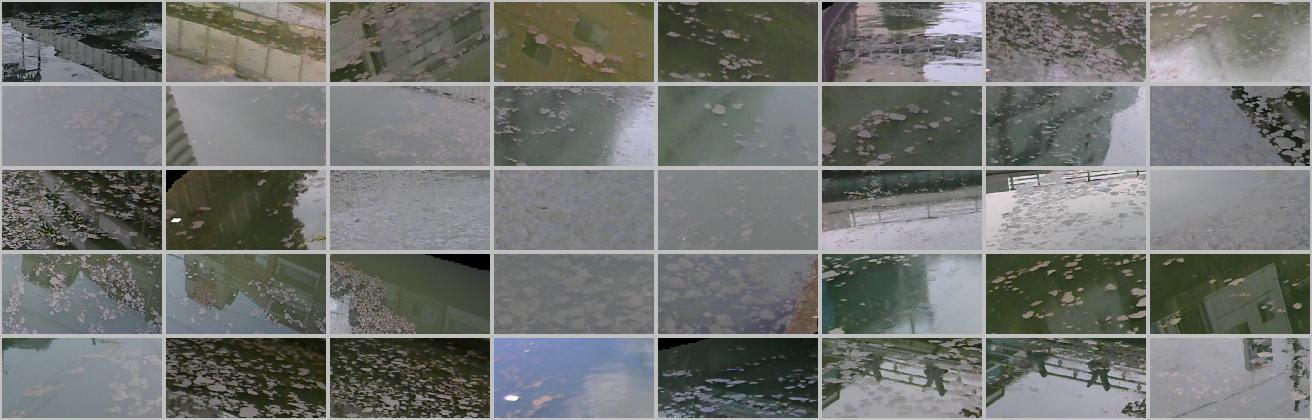}
\end{center}
\caption{Example of Class \textbf{``C1 grow-thick scum"} Patch Images}
\label{fig-m2C1}
\end{figure}

Furthermore, as illustrated in \textbf{Fig.15}, we show the C2 class of images for classification training with 4,308 images. We randomly sampled 40 patch images from the C2 background class of the dataset. The C2 class was the domain where all of images contained background concrete structures, steel pipes and trees. Furthermore, there have been mixed images between zero scum on the river and those structures under the bridge that are in some dark location. 
\begin{figure}[h]
\begin{center}
\includegraphics [width = 88mm] {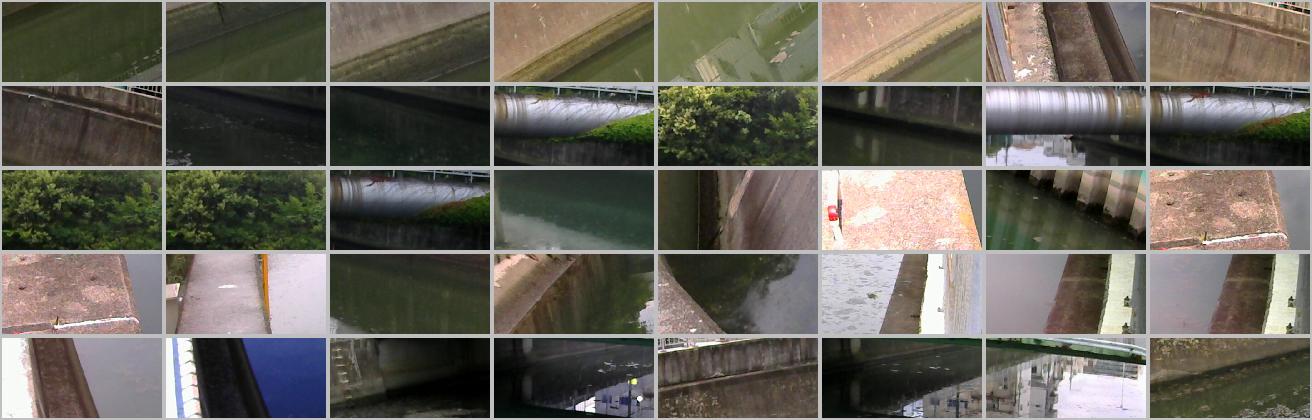}
\end{center}
\caption{Example of Class \textbf{``C2 background"} Patch Images}
\label{fig-m3C2}
\end{figure}

\section*{Acknowledgment}
We thank Takuji Fukumoto and Shinichi Kuramoto (MathWorks Japan) who contributed to supporting the MATLAB resources on Data Augmentation for Deep Learning.

\ifCLASSOPTIONcaptionsoff
  \newpage
\fi



%

%

\begin{IEEEbiographynophoto}{Takato Yasuno}
His ORCID number is 0000-0002-4796-518X. He received his D.E. (Doctor of Engineering) degree from Tottori University. He has over 18 years of experience as a consulting engineer in infrastructure asset management. He works at Yachiyo Engineering Co., Ltd as a senior researcher, developing data mining processes and providing AI practice advisory. He has interests in Machine Learning, Computer Vision and Pattern Recognition, specifically Predictive Diagnosis/Deterioration Prognosis and Sustainable Urban Environment. He has 14 accepted papers and 12 arXiv preprints. He is a member of JSAI. 
\end{IEEEbiographynophoto}

\begin{IEEEbiographynophoto}{Junichiro Fujii}
He received his B.E. degree from Kyoto University and M.A.S.(Interdisciplinary Information Studies) degree from University of Tokyo. He has over 15 years of experience in information systems development, and presently works as a researcher at Yachiyo Engineering Co., Ltd. His research interest is applying artificial intelligence to the field of civil engineering.
\end{IEEEbiographynophoto}

\begin{IEEEbiographynophoto}{Masazumi Amakata}
He received his B.E. degree from Kyoto University and D.E.(Doctor of Engineering) degree from Kanazawa University. He has over 20 years of experience in flood control, water utilization, river environment and is a specialist in fluid analysis. He works as a director of the research institute in Yachiyo Engineering Co., Ltd. His research interest involves applying Machine Learning and Deep Learning to the field of civil engineering.
\end{IEEEbiographynophoto}



\vfill


\end{document}